\documentclass[11pt, a4paper]{article} 

\usepackage[german,english]{babel} 
\usepackage{booktabs} 
\usepackage{comment} 
\usepackage[utf8]{inputenc} 
\usepackage[T1]{fontenc} 

\usepackage{amsmath,amsfonts,amsthm} 
\usepackage{tikz} 
\usepackage{tikz-cd}
\usepackage{hyperref}
\usetikzlibrary{positioning,arrows} 
\usetikzlibrary{decorations.pathreplacing} 
\usepackage{tikzsymbols} 
\usepackage{blindtext} 

\usepackage{geometry} 

\usepackage{setspace} 
\usepackage[T1]{fontenc} 
\usepackage[utf8]{inputenc} 

\usepackage{XCharter} 

\usepackage{fancyhdr} 
\usepackage{natbib} 
\usepackage{har2nat} 

\title{Training spiking neural networks for reinforcement learning} 
\author{
Sneha Aenugu 
  }

\date{\small \today} 

\begin{document}


\maketitle 


\setcounter{page}{1} 

\section*{Abstract}
Neurons in the brain communicate with each other through discrete action spikes as opposed to continuous signal transmission in artificial neural networks. Therefore, the traditional techniques for optimization of parameters in neural networks which rely on the assumption of differentiability of activation functions are no longer applicable to modeling the learning processes in the brain. In this project, we propose biologically-plausible alternatives to backpropagation to facilitate the training of spiking neural networks. We primarily focus on investigating the candidacy of reinforcement learning (RL) rules in solving the spatial and temporal credit assignment problems to enable decision-making in complex tasks. In one approach, we consider each neuron in a multi-layer neural network as an independent RL agent forming a different representation of the feature space while the network as a whole forms the representation of the complex policy to solve the task at hand. In other approach, we apply the reparametrization trick to enable differentiation through stochastic transformations in spiking neural networks. We compare and contrast the two approaches by applying them to traditional RL domains such as gridworld, cartpole and mountain car. Further we also suggest variations and enhancements to enable future research in this area.

\section*{Introduction}

Most reinforcement learning algorithms (RL) primarily fall into one of the two categories, value function based and policy based algorithms. The former category of algorithms such as Qlearning\cite{qlearning} and SARSA, express value function as a mapping from a state space to a real value, which indicates how good it is for an agent to be in a state. Policy based algorithms such as REINFORCE\cite{williams}, on the other hand, learn policy as a mapping from state space to action space, which tells the agent the best action in each state to maximize its reward. Both types of algorithms, thus, need to address the problem of learning a representation of the state space to solve the task at hand. As the task gets more complex, so does the representation to be learned. 

Deep neural networks are a viable means to learn these mappings due to the potential for rich feature representations afforded by the hierarchical structure of these networks. Deep Q networks \cite{deeprl} trained with backpropagation algorithm proved successful in solving complex RL domains. Successful though they are in learning complex representations for certain RL tasks, these optimization techniques  suffer from several drawbacks. Some of them are
\begin{enumerate}
    \item Deep networks require large amounts of data for optimizing their parameters compared to biological networks which are able to generalize using fewer data samples
    \item Biological networks are energy-efficient whereas training deep networks consume a lot of power.
    \item Deep networks are not robust to adversarial attacks whereas biological networks are inherently good at dealing with noisy/incomplete data and therefore are less prone to adversarial attacks.
\end{enumerate}

In this context, it is worthwhile to investigate optimization process in biological networks to gain insights into the functioning of the human brain which can help design robust intelligent machines. Spiking neural networks which are close approximations to biological networks are built on spiking neurons which emulate most of the features of the neurons in the brain.  

Besides biological plausibility, a spiking neuron holds a tremendous computational potential that is yet to be harnessed. The variability in the spiking pattern of a single neuron can be attributed to a rich set of factors - the stimulus, its past spiking record, the excitatory/ inhibitory effects from its neighbours. Further in the context of temporal coding, where the precise pattern of spikes rather than just their density, stores relevant information, this single computational unit is capable of representing complex feature maps. Incorporating a spiking neuron in the network does not support the traditional optimization techniques to train artificial neural networks. Credit assignment through backpropagation requires the activation function of the neuron to be continuous and differentiable. As spiking neuron's activation function does not meet the requirements of continuity and differentiability, alternative techniques need to be developed to optimize the parameters of spiking neural networks which is the central theme of this project.

In this project, we analyze two different approaches to train spiking neural networks to perform reinforcement learning tasks such as maze navigation and cartpole balancing. The approaches are
\begin{enumerate}
    \item We build on the approach of \cite{pgcn} and propose a multi-agent framework where each spiking neuron acts an independent RL agent which optimizes its firing policy to maximize the reward it receives. Each neuron in the network updates its policy parameters using local information from the neighbouring neurons and global reward received from the environment.
    \item We assume each spiking neuron samples its actions from a firing policy thus forming a stochastic node in the network. We use the reparameterization trick to enable differentiation through a stochastic node thus enabling backpropagation through the network.
\end{enumerate}

We derive the learning rules for the above described approaches and apply them to optimize the parameters in a multi-layered spiking neural network to solve gridworld, cartpole and mountaincar problems. In sections 1 through 6, we discuss the multi-agent actor-critic framework developed on the constraints of biological plausibility. Section 7 discusses the reparameterization trick and how to use backpropagation to train spiking neural networks. In section 8, we describe the several case studies studied to validate our claims. We discuss the findings and conclude in section 9.

\section{Reinforcement learning with a network of spiking agents}
Neuroscientific theory (\cite{dopamine}) indicates that critic dopaminergic neurons fire in response to the reward prediction error and these reinforcement signals which are broadcast globally are believed to guide learning in actor neurons throughout the frontal cortex and basal ganglia. We build on this theory to propose a multi-agent actor critic framework based on the theory of policy gradient coagent networks \cite{pgcn}. In this framework, we let a multi-layered neural network describe a complex policy to solve a given RL task where each neuron in the network acts an independent RL agent whose policy forms a feature representation of part of the state space. The neurons in the higher layers use the feature repesentations of the neurons in the lower layer to form more complex representations of the state space, all the while ensuring the learning algorithm and information transmission is strictly local thus satisfying the biological-plausibility constraints. 

\section{Related Work}

\subsubsection{Hedonism}
The concept of a neuron as an entity which optimizes its activity to maximize its reward dates back to the formulation of hedonistic neuron by \cite{hedonistic}. A similar formulation by \cite{asn} introduced an associative memory system called an associative search network containing neuron-like adaptive elements and a predictor where the predictor sends reward signals to each of the adaptive elements which independently optimizes its parameters. \cite{seung} formulated hedonistic synapses where connections between the neurons are modeled after chemical synaptic transmission. The probability of vesicle release or failure for a synapse is modulated by a global reward.

\subsubsection{Learning by reinforcement in spiking neural networks}
\cite{florian} applied reinforcement learning algorithm to a stochastic spike response model of spiking neuron to derive learning rules for reward modulated spike-timing dependent learning. \cite{xie} formulated a learning rule correlating irregular spiking in a network of noisy integrate-and-fire neurons with the global reward and showed that it performs a stochastic gradient descent on the expected reward. \cite{glm} employed a GLM model of neurons with first-to-spike coding to learn policies for neuromorphic control. 
\subsubsection{Multi-agent learning}
\cite{kalman} trains multiple RL agents using a global reward signal where each agent models the contribution of unseen agents as an additive noise process that can be estimated through kalman filtering.

\section{Background, Preliminaries and Notation}
A reinforcement learning (RL) domain expressed as a Markov Decision Process (MDP) is defined by the state space, $\mathcal S$, the action space $\mathcal A$, a state transition matrix, $\mathcal P: \mathcal S \times \mathcal A \rightarrow \mathcal S$ and a reward function, $\mathcal R: \mathcal S \times \mathcal A \rightarrow \mathbb{R}$. A policy is a distribution of action probabilities conditioned on state space and is defined as $\pi(s,a,\theta)=\Pr(A_t=a|S_t=s)$ where $\theta$ denotes the parameters of the policy. A state-value function of the policy $\pi$ is defined as the expected return, $V^{\pi}(s) =\mathbb{E}[G |S_0=s,\pi]$ where $\gamma$ is the discount factor and $G=\sum_{t=0}^\infty \gamma^t R_{t}$ is the discounted return.
\subsection{TD($\lambda$)}
Temporal difference learning (TD) algorithms are used to evaluate a policy $\pi$ by learning the state value function $V_\pi^\theta$. TD(0) algorithm updates its parameters $\theta$ based on the difference in two successive predictions (TD error) of $V_\pi^\theta$. The TD error is defined as $\delta_t = R_{t} + \gamma V_\pi^\theta(S_{t+1}) - V_\pi^\theta(S_t) $. TD(0) update is given by $\theta \leftarrow \theta + \alpha \delta_t \frac{\partial V_\pi^\theta(S_t)}{\partial \theta}$. TD($\lambda$) is an extension of TD(0) algorithm where the contributions of all the states prior to the current reward are taken into account weighted by eligibility, $e_v \leftarrow \gamma \lambda e_v + \frac{\partial V_\pi^\theta(S_t)}{\partial \theta}$ and updated as  $\theta \leftarrow \theta + \alpha \delta_t e_v$.
\subsection{Policy gradient algorithms}
Policy gradient class of algorithms formulate the policy as a complex mapping from state space to action space through function approximation. The optimal parameters for the policy are estimated by descending the gradient of the expected discounted return, $\theta \leftarrow \theta + \alpha \nabla J(\theta)$ where $J(\theta) =\mathbb{E}[G |\theta]$. In REINFORCE, the value of the gradient, $\nabla J(\theta)$ is estimated as $\nabla J(\theta) \propto \mathbb{E_\pi}\left[G_t \frac{\partial \ln\pi(S_t,A_t,\theta)}{\partial \theta}\middle|\theta\right]$. The high variance exhibited by the REINFORCE algorithm can be mitigated by incorporating the estimate of state-value function as a baseline. The gradient with baseline is given as $\nabla J(\theta) \propto \mathbb{E_\pi}\left[(G_t - V(S_t)) \frac{\partial \ln\pi(S_t,A_t,\theta)}{\partial \theta}\middle|\theta\right]$.
\subsection{Actor-critics}
Actor-critics belong to the class of policy gradient algorithms where the policy and the state-value function are learned in parallel by an actor and a critic respectively. The gradient $\nabla J(\theta)$ in actor-critics replaces the $G_t$ in REINFORCE with a one step return $R_t+V(S_{t+1})$ forming the update equation 
\begin{equation}
\nabla J(\theta)\propto \mathbb{E_\pi}\left[\delta_t \frac{\partial \ln\pi(S_t,A_t,\theta)}{\partial \theta}\middle|\theta\right]
\end{equation}
where $\delta_t = R_t + V(S_{t+1}-V(S_t) $ is the TD error. Critic estimates the state-value function using a TD algorithm. Although advanced actor-critic methods estimate state-value function by performing the gradient descent on the expected discounted return.

\subsection{Conjugate Markov Decision Processes}
\cite{comdp} developed a multi-agent learning framework where a set of coagents work towards discerning the underlying structure in feature space which is to be used by the  agent to solve the original MDP. Thus the problem of identifying a mapping from the state space to an action space can be broken down into sub tasks which can be delegated to each of the coagents. Each of these sub tasks can in turn be modeled as an MDP and thus referred to as conjugate markov decision processes (CoMDPs). Let $\mathcal S$ be the state space of the original MDP. The state space of each coagent, $\mathcal S_c$, is a subset of $\mathcal S$ and its action space is $\mathcal A_c$. The state space of the agent, $\mathcal S_A = \{\mathcal S, \mathcal A_{c_1}\,..,\mathcal A_{c_n}\}$, is now extended to include the action space of the coagents and the action space of the agent $\mathcal A$ is the action space of the original MDP. Let $\theta=(\theta_A, \theta_{c_1},..,\theta_{c_n})$ be the parameters of the MDP where $\theta_A$ are the parameters of the agents and $\theta_{c_i}$ are the parameters for the coagent $i$. In such a formulation it can be shown that descending the policy gradient on the MDP as a whole is equivalent to descending the policy gradients on each of the CoMDPs separately. 
\begin{equation}
    \nabla_{\theta} J(\theta) = (\nabla_{\theta_A} J_{\theta}, \nabla_{\theta_{c_1}} J(\theta),..,\nabla_{\theta_{c_n}} J(\theta))
\end{equation}
Thus the primary conclusion of the coagent theory can be summarized as follows: \emph{In a coagent network, optimizing the policy of each of the coagents separately is equivalent to optimizing the policy of the MDP as a whole}.  
\subsection{Policy Gradient  Coagent Networks}
\cite{pgcn} introduced a class of actor-critic algorithms to optimize the performance of a modular coagent networks in solving an RL task. The coagent network, termed as policy gradient coagent network (PGCN), consists of a set of coagents each optimizing its own policy by descending its local policy gradient modulated by the TD error delivered by the global critic. This is reminiscent of dopaminergic neurons broadcasting reward signals to a population of neurons to modulate their synaptic plasticity\cite{dopamine}. 

\begin{figure}[htpb]
\centering
\begin{tikzpicture}
  \node (rect) (1) at (0,2) [draw,thick,minimum width=1.7cm,minimum height=1cm,text width=1.7cm,align=center] {Critic};
  \node (rect) (2) at (0,0) [draw,thick,minimum width=1.7cm,minimum height=1cm,text width=1.7cm,align=center] {$A$\\$\pi_a(s_a,a,\theta_a)$};
  \node (rect) (3) at (-2,-2) [draw,thick,minimum width=2.7cm,minimum height=1cm,text width=2cm,align=center] {\small $C_1$\\$\pi_{c_1}(s_{c_1},a_{c_1},\theta_{c_1})$};
  \node (rect) (4) at (2,-2) [draw,thick,minimum width=2.7cm,minimum height=1cm,text width=2cm,align=center] {\small$C_2$\\$\pi_{c_2}(s_{c_2},a_{c_2},\theta_{c_2})$};
  \path [<-,>=stealth](3) edge[bend left] node [left] {$\delta$} (1);
  \path [<-,>=stealth](4) edge[bend right] node [right] {$\delta$} (1);
  \path [<-,>=stealth](2) edge[bend right] node [right] {$\delta$} (1);
  \path [<-,>=stealth](1) edge[bend right] node [right] {$a$} (2);
  \path [<-,>=stealth](2) edge[bend right] node [right] {$a_{c_1}$} (3);
  \path [<-,>=stealth](2) edge[bend left] node [left] {$a_{c_2}$} (4);
\end{tikzpicture}

\caption{A policy gradient coagent network with agent $A$, coagents $C_1$, $C_2$ and the global critic. }
\end{figure}
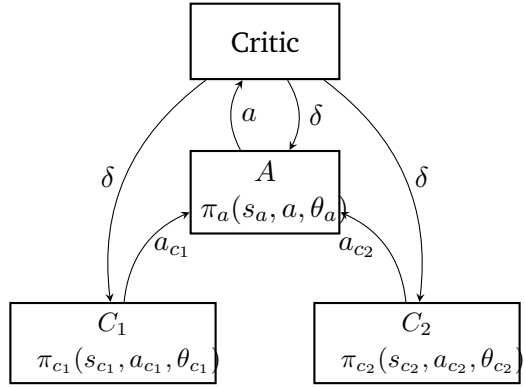

Consider the PGCN shown in the Figure 1. The policy of the coagent $c$ is defined as $\pi_c(S_c, A_c, \theta_c)$. Each of the coagents executes its policy $\pi_c$ and the agent $A$ takes the actions of the coagents into account and executes its policy $\pi_a(S_a, A, \theta_a)$ which is the action of the MDP at that instant. A global critic, evaluates the action in the environment and returns a TD error $\delta$ which is then delivered to each of the coagents. The coagent, c, then optimizes its parameters through stochastic gradient descent as per Equation (1) as follows
\begin{equation}
    \theta_c = \theta_c + \alpha_c \delta \frac{\partial \ln\pi_c(S_c,A_c,\theta_c)}{\partial \theta_c}
\end{equation}
The global critic, $C$, uses the same TD error to update its estimate of state-value function using the TD($\lambda$) algorithm. 
The coagent framework, thus facilitates a biologically plausible learning rule. A spiking neuron can be modeled as a coagent where its stochastic firing pattern can be defined as its policy. We now proceed to describe the mathematical model of the spiking neuron which can be modeled as a coagent.

\section{Designing the neural agent: spiking neuron models}

Spike trains of neurons \emph{in vivo} are highly irregular and not reproducible for an identical constant stimulus. This variability in neuronal spike trains is posited to be conducive to learning. \cite{mass} showed that networks of noisy spiking neurons can simulate in real-time the functioning of any McCulloch-Pitts neuron/ multilayer perceptron. The unreliable nature of the spiking process makes it feasible for the neuron to be modeled as an RL agent.

\subsection{Memoryless Ising model of spiking neuron}

In this section we consider an energy-based model (Ising) of firing activity of a population of spiking neurons \cite{optimal}. Consider a network of $N$ neurons whose spike trains are discretized into bins of width $\Delta t$ where for any neuron $i$, $\sigma_i(t) = 1$, if the neuron fired in time bin t and $\sigma_i(t) = -1$ when the neuron is inactive. Thus at any instant of time $t$, an $N$-bit representation of the network activity is formed by the firing pattern of the groups of neurons. The joint probability distribution of the spiking activity is given by the Boltzmann distribution given below.
\begin{equation}
  \Pr(\sigma(t)) = \frac{1}{Z(b_i, W_{ij})}\exp\left(\sum_i b_i \sigma_i(t) + \sum_{i,j} W_{ij} \sigma_i(t) \sigma_j(t)\right)
\end{equation}

where $Z(b_i, W_{ij})$ is the partition function. The conditional probability of spiking of a single neuron given the activity of its neighbours is given by

\begin{equation}
  \Pr(\sigma_i = 1 | s) = \frac{\exp(A(t))}{\exp(A(t)) + \exp(-A(t))}
\end{equation}
where $A(t) = b_i + \sum_j W_{ij} \sigma_j$. 

The firing policy of the neuron is thus defined by the above conditional probability with parameters as $b_i$, the inherent tendency of the neuron to fire and $W_{ij}$, the strength of synaptic connections with its neighbours. The form of representation is similar to that of Restricted Boltzmann Machines (RBM). RBM uses contrastive divergence algorithm as a learning rule to update its parameters instead here we propose to adjust the synaptic weights using RL updates. This formulation of spiking neuron is defined as an RL agent where the set of actions available to the agent are to fire ($a=1$) or not to fire ($a=-1$). The policy of the neuron is defined by $\pi(s,a) = \Pr(a_t=a|s_t=s) = \Pr(\sigma_i=1|s)$. 	

Further gradation of neuronal spiking activities can be incorporated into the policy rather than just fire/ silent. If the action $a=0$ represents firing at mean firing rate, $a=1$, $a=2$ can represent firing at one, two standard deviations above the mean activity where $a=-1$, $a=-2$ can present firing below the mean level. Then the policy of the neuron can be represented as 

\begin{equation}
  \pi(s,a=a_k) = \frac{\exp(a_kA(t))}{\sum_{-2}^2\exp(a_iA(t))}
\end{equation}
where $A(t) = b_i + \sum_j W_{ij} \sigma_j$. This form of parameterization can be used for categorical representation of firing activities of spiking neurons.

\subsubsection{PGCN update emulates Hebbian/Anti-Hebbian learning in Ising model of spiking neurons}

\begin{proof}

  Spike trains from a neuron are discretized into small time bins of duration $\Delta t$. Then $x_i(t)=1$ indicates firing of the neuron and $x_i(t)=-1$ indicates silence.
  The joint probability of the firing patterns of a network of neurons is approximated by the Ising model\cite{gasper}. As the stimulus/input pattern is fixed, the firing probability of the neuron connected to $K$ neurons of the previous layer is given by

  \begin{equation}
    Pr({x_i} | s) =   \frac{\exp\left(b_ix_i + \sum_{k=1}^K W_{ki} x_i x_k\right)}{Z}
  \end{equation}

  \begin{equation}
    Pr(x_i=1 | s) =   \frac{\exp\left(b_i + \sum_{k=1}^K W_{ki} x_k\right)}{\exp\left(b_i + \sum_{k=1}^K W_{ki} x_k\right) + \exp\left(-b_i - \sum_{k=1}^K W_{ki} x_k\right)}
  \end{equation}

  Let $u = b_ix_i + \sum_{k=1}^K W_{ki} x_i x_k$,

  \[
    \pi_i(s,a)= \begin{cases}
      \frac{\exp(u)}{\exp(u) + \exp(-u)}, & \text{if}\ a = 1\\
   \frac{\exp(-u)}{\exp(u) + \exp(-u)}, & \text{if}\ a= -1 \end{cases}
\]

  \[
    \frac{d\ln\pi_i(s,a)}{dW_{ki}}= \begin{cases}
      \frac{2 \exp(-u)}{\exp(u) + \exp(-u)}x_k, & \text{if}\ a = 1\\
   \frac{-2 \exp(u)}{\exp(u) + \exp(-u)}x_k, & \text{if}\ a= -1 \end{cases}
\]

According to the policy gradient theorem, weight update is given by

\begin{equation}
  \Delta W_{ki} = \alpha\delta_{TD}\frac{d\ln\pi_i(s,a)}{dW_{ki}}
\end{equation}

  \[
    \Delta W_{ki}= \begin{cases}
      \frac{2\exp(-u)}{\exp(u) + \exp(-u)}\alpha\delta_{TD} x_k, & \text{if}\ a = 1\\
    -\frac{2\exp(u)}{\exp(u) + \exp(-u)}\alpha\delta_{TD}x_k, & \text{if}\ a= -1 \end{cases}
\]

From the above equation we can see that,  when $\delta_{TD} > 0$, the weight update follows hebbian learning (if $x_k = 1$ and $a=1$, the update is positive and vice versa) and when $\delta_{TD} < 0$, the weight update follows anti-hebbian learning.
\end{proof}

\subsection{Stochastic Leaky Integrate \& Fire Neuron}

In this section, we consider a spiking neuron model (\cite{spikingneuron}) which maintains a decaying memory of past inputs in its membrane potential. 

Consider a post-synaptic neuron that receives inputs from multiple pre-synaptic neurons through their respective synaptic connections. The membrane potential of a post-synaptic neuron due to the spike trains in the presynaptic neurons at any instant of time is given by the equation

\begin{equation}
  u(t) = b + \sum_{k=1}^K \sum_{i=1}^N W_{k}z_i(t - t_i)
\end{equation}

where $b$ is the bias of the neuron to fire, $W_k$ is the synaptic weight to the pre-syanptic (input) neuron $k$ and $z_i$ is the post-synaptic action potential caused in the neuron due to a spike $i$ occuring at time $t_i$ in the pre-synaptic neuron $k$.

\subsubsection{PGCN update emulates STDP learning in Leaky-Integrate-Fire model of spiking neurons}

\begin{proof}
  Here we consider the time-to-first-spike neural coding. We assume the relevant information is encoded in the pattern of neurons which fired first among a group of neurons. Consider 3 spiking agents which are laterally inhibited by each other. When one of them fires, the episode terminates for all the 3 agents.

The probability(policy) of the neuron firing at any instant is given by

\begin{equation}
  \pi(a=1,u,t)  = \Pr(a=1 | u) = \sigma(u(t) - \theta)
\end{equation}
\begin{equation}
  \pi(a=0,u,t) = 1 - \sigma(u(t) - \theta)
\end{equation}

When one of the neurons fire, the episode ends for all the three neurons and the critic gives a TD error of $\delta_{TD}$.

The weight update is given by the following policy gradient equation.

\begin{equation}
  \Delta W_k = \alpha\delta_{TD}\frac{d\ln(\pi(u,a))}{dW_k}
\end{equation}

\[
   \frac{d\ln(\pi(u,a))}{dW_k}= \begin{cases}
     \frac{\sigma'(u(t) - \theta)}{\sigma(u(t) - \theta)}\sum_{i=1}^N z_i(t - t_i), & \text{if}\ a = 1\\
   -\frac{\sigma'(u(t) - \theta)}{(1 - \sigma(u(t) - \theta))}\sum_{i=1}^N z_i(t - t_i), & \text{if}\ a= 0 \end{cases}
\]

In a Leaky-Integrate-Fire neuron, the post-synaptic potential induced in a neuron by a spike in a pre-synaptic neuron decays exponentially from the time of the spike.
\begin{equation}
  z_i(t) = \exp((t - t_i)/\tau)
\end{equation}

\[
   \Delta W_k= \begin{cases}
     \alpha\delta_{TD} \frac{\sigma'(u(t) - \theta)}{\sigma(u(t) - \theta)}\sum_{i=1}^N \exp((t - t_i)/\tau), & \text{if}\ a = 1\\
 -\alpha\delta_{TD}\frac{\sigma'(u(t) - \theta)}{(1 - \sigma(u(t) - \theta))}\sum_{i=1}^N \exp((t - t_i)/\tau), & \text{if}\ a= 0 \end{cases}
\]

The above update is reminiscent of Spike-Timing-Dependent-Plasticity learning rule where the synaptic updates are depending on the relative timing of the pre-post synaptic spikes. The neuron which fired at time $t$ will have a positive update and the neuron which has failed to fire will have a negative update. Inhibition is not modeled explicitly here as it is treated as a signal for termination of the episode.

\end{proof}

\subsection{Generalized Linear Model of a Spiking Neuron}
A Generalized Linear Model (GLM) of a spiking neuron \cite{pillow,truccolo} is a computationally tractable framework designed to capture the spatio-temporal correlations of the neuron's spiking pattern with that of its sensory/stimulus representation and the activity of the neighbouring neurons. In this model, the firing activity of the neuron is parameterized by a set of linear filters each attributing the spike train variability to a biologically realistic phenomenon. The stimulus filter or a spatio-temporal receptive field, akin to the filters in convolutional neural networks, converts the stimulus into the relevant higher dimensional representation. The post-spike filter relates the influence of the spike history dynamics such as refractoriness and bursting on the current firing activity of the neuron. Inhibitory/ excitatory effect of the firing activity of the neighbouring neurons is captured by the coupling filters. The pictorial representation of the GLM spiking neuron is shown in Figure 2 \cite{glmfig}.
\begin{figure}[htpb]
\centering
\includegraphics[width=150mm]{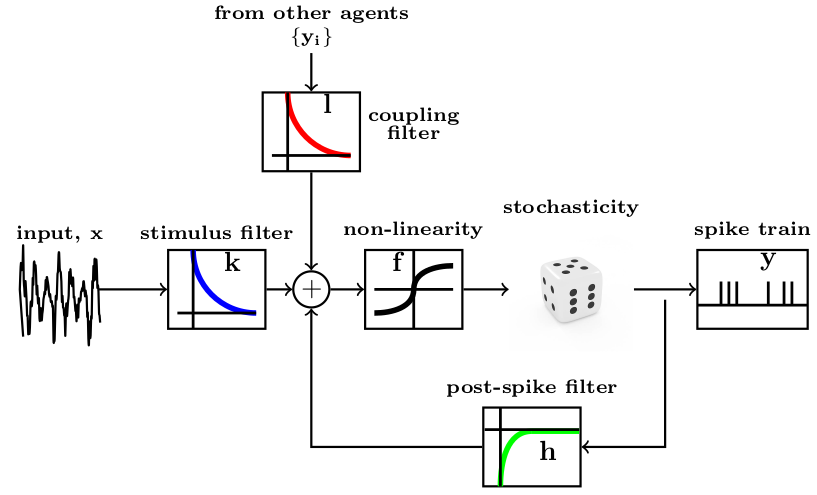}
\caption{A generalized linear model(GLM) of the spiking neuron}
\end{figure}

The conditional spiking activity of the neuron at a given time $t$ is assumed to be sampled from an exponential family distribution whose expected value, $\lambda(t)$, is related to the linear combination of its filter responses - $\mathbf{k}$ (stimulus filter), $\mathbf{h}$ (post-spike filter), $\{\mathbf{l_i}\}$ (coupling filters) and its baseline firing rate $\eta$, through a link function $f$. 
\begin{equation}
    f(\lambda(t)) = \mathbf{k}\cdot \mathbf{x} + \mathbf{h}\cdot \mathbf{\zeta} + (\sum_i  \mathbf{l_i}\cdot \mathbf{\xi_i} ) + \eta 
\end{equation}

where $\mathbf{x}$ is the spatio-temporal stimulus pattern, $\mathbf{\zeta}$ is the spike history of the neuron and $\mathbf{\xi_i}$ is the firing activity of the $i^{\textnormal{th}}$ neighbouring neuron. Here $f$ is  an invertible function. If $f^{-1}$ is an exponential function, $\lambda(t)$ can represent conditional spike rate, whereas if $f^{-1}$ is sigmoidal, $\lambda(t)$ can represent conditional probability of firing at any instant of time. The stochasticity observed in the spike trains \emph{in vivo} is accounted for by the stochastic spiking module as shown in Figure 2 which could either incorporate Poisson/Bernoulli randomness depending on whether $\lambda(t)$ represents instantaneous firing rate/ firing probability. 

The optimal parameters of the above model ($\theta^*=(\mathbf{k},\mathbf{h},\{\mathbf{l_i}\}) $) can be estimated by the maximum likelihood of observing the spike train response of the neuron given the stimulus parameters and activity of neighbouring neurons. $\theta^*$ can be estimated by computing the gradient and hessian of the log likelihood of observing a given spike train response. 

In this study, we formulate the GLM spiking neuron as a coagent in the framework of PGCN where the instantaneous firing rate/ firing probability is its policy and with the set of GLM filters as its policy parameters. The spiking coagent can optimize its policy parameters by descending the policy gradient with respect to its parameters. We now discuss the proposed learning rule to train a network of spiking coagents to solve a reinforcement learning task.

\subsection{Linear-nonlinear-Poisson cascade neuron model}

The spiking coagent with GLM neuron model attributes its spiking policy to a rich set of factors - stimulus, spiking history, interactions with neighbouring coagents. One obvious simplification of the model is to limit the set of factors by reducing the GLM neuron model to the Linear-nonlinear-Poisson cascade neuron model which can be described the equations below.
\begin{equation}
    \lambda(t) = f(\mathbf{k} \cdot \mathbf{x}) \quad y(t)|x(t) \sim \text{Poiss}(\lambda(t))
\end{equation}
where $\lambda(t)$ is the conditional firing rate, $\mathbf{k}$ is the stimulus filter and $y(t)$ is the spike train response of the coagent. This formulation is suitable where we need to use rate coding of instead of temporal coding for encoding the response of the coagents. The action of the neuron, ita instantaneous firing rate, can be executed by a gaussian policy with $\lambda(t)$ as the mean as shown below.
\begin{equation}
    \pi_c(s_c, a_c=y, \theta_c=\{\mathbf{k}, \sigma\}) =
    \frac{1}{(2\pi)^{\frac{1}{2}}\sigma}e^{-\frac{y-\lambda(t)}{2\sigma^2}}
\end{equation}

The continuous policy representation can thus extend the coagents to take continuous actions rather than discrete spikes. We can further simplify the model by reducing the action of the coagent to a single spike rather than spike train by replacing the Poisson sampling with Bernoulli sampling. The policy can now be represented by 
\begin{equation}
    \pi_c(s_c, a_c=y, \theta) = \lambda(t)^{[y = 1]}  (1 - \lambda(t))^{[y=-1]}
\end{equation}
The weight updates for the coagent for the above policy are
\begin{equation}
    \Delta \theta =  \alpha \Big((1-\lambda(t))^{[y=1]} + (-\lambda(t))^{[y=-1]}\Big) x
\end{equation}
In the case of single spikes, the update rule closely resembles Hebbian learning as can be seen from Equation(15) where if $x$ and $y$ have the same sign, the parameter is increased and vice versa.

\section{PGCN learning rule for a network of GLM spiking agents}
Consider a complex state-action representation in a reinforcement learning domain formed by a deep hierarchical network of GLM spiking coagents. We now describe how to represent a policy using a network of spiking coagents to solve a reinforcement learning task and derive the corresponding learning rules to update the policy parameters under the PGCN framework. 

\subsection{Architecture and encoding}
We will first discuss the encoding of the state space in terms of a spatio-temporal spiking pattern. Each state can be represented by $S \times K$ vector of binary values (1/-1) where $S$ is the spatial factor and $K$ is the temporal factor. Here $S$ can be interpreted as number of neurons encoding the stimulus and $K$ as the length of the spike trains (measured as discrete time bins) considered for each of the $S$ neurons.  Any complex state representation can be encoded by conversion into a spatio-temporal pattern. In case the state involves a grayscale/color image, the state can be represented by instantaneous firing rate instead of firing probability. However in this study, we primarily consider the spike train as a series of binary spikes. In extensions, we discuss how to deal with spike trains as a series of instantaneous firing rates. 

The state neurons form the first layer of the deep coagent network which is the stimulus for the first coagent layer. The rest of the network is hierarchically organized with the spike train response of one layer of coagents forming the stimulus for the succeeding layer of coagents. Coagents within the same layer can have coupling connections with excitatory/inhibitory effects as discussed in the earlier section. 

Consider a spiking coagent, C, in a layer (n) of the network. Let $\mathbf{x_t} = (\mathbf{x^{(0)}},..,\mathbf{x^{(S)}})$ be the stimulus of the coagent C, where $\mathbf{x^{(i)}}=(x_{0}^{(i)},..,x_{t-1}^{(i)})$ is the temporal spike response pattern of the ith coagent of the preceding layer (n-1). Let $\mathbf{\zeta_t}=(y_{0},..,y_{t-1})$ be a vector of spike history of the coagent. Let $\mathbf{\xi_{t-1}} = (\xi_{t-1}^{(0)},..,\xi_{t-1}^{(n)})$ be the firing activity of the neurons in the same layer at a previous time instant assuming excitatory/inhibitory effects of the activity of the neighbouring neurons kick in during the next time instant thereby avoiding recurrence in the updates. The state space of the coagent C is given by the tuple $(\mathbf{x_t},\mathbf{\zeta_t},\mathbf{\xi_{t-1}})$. The conditional firing probability of a coagent at a time $t$ is given by
\begin{align}\nonumber
    \text{logit}(\lambda(t))&= \text{logit}(\Pr \left(y_t=1) \middle|\mathbf{S_t}=(\mathbf{x_t},\mathbf{\zeta_t},\mathbf{\xi_{t-1}}) \right)  \\
    &= \mathbf{k}\cdot \mathbf{x_t} + \mathbf{h}\cdot \mathbf{\zeta_t} +  \mathbf{l}\cdot \mathbf{\xi_t}  + \eta 
\end{align}

where$\mathbf{k}, \mathbf{h}, \mathbf{l}$ are filters of the coagent C and the link function $f$ in Equation (1) is chosen to be the logit function where $\text{logit}(p) = \log\left(\frac{p}{1-p}\right)$. 
\begin{align}\nonumber
   \Pr \left(y_t=1) \middle|\mathbf{S_t}=(\mathbf{x_t},\mathbf{\zeta_t},\mathbf{\xi_{t-1}}) \right) = \sigma(\mathbf{k}\cdot \mathbf{x_t} + \mathbf{h}\cdot \mathbf{\zeta_t} + \mathbf{l}\cdot \mathbf{\xi_t} + \eta)
\end{align}
where $\sigma$ is the sigmoid/logistic function.

\subsection{Learning updates}
Consider a single MDP time step. The conditional probability of observing a spike train $\mathbf{y}$ in that time step is given by
\begin{align}
   \Pr \left(\mathbf{y_\tau}=\mathbf{y}) \middle|\mathbf{S_\tau}=(\mathbf{x},\mathbf{\zeta},\mathbf{\xi}) \right)  = \prod_{t\in t_s} \lambda^{(\tau)}(t) \prod_{t \in t_{ns}}(1-\lambda^{(\tau)}(t))
\end{align}
where $\lambda(t) = \sigma(\mathbf{k}\cdot \mathbf{x_t} + \mathbf{h}\cdot \mathbf{\zeta_t} + \mathbf{l}\cdot \mathbf{\xi_t} + \eta) $ and $\tau$ is the time on the MDP scale. $t_s$ denotes times at which spike occurs and $t_{ns}$ denotes the times at which there is no spike. $(\mathbf{x},\mathbf{\zeta},\mathbf{\xi})$ denote the complete spatio-temporal spiking patterns expressed in the time scale of the MDP.
Equation (7) represents the policy parametrization of the coagent c, $\pi_c(\mathbf{s}, \mathbf{a}, \theta)$, where its extended action is the spike train response $\mathbf{a_c} = \mathbf{y}$, the state space $\mathbf{s_c}$ is the tuple $\mathbf{s_c}=(\mathbf{x_t},\mathbf{\zeta_t},\mathbf{\xi_{t-1}})$ and the parameter vector $\theta_c = (\mathbf{k},\mathbf{h},\mathbf{l})$. \begin{equation}
    \pi(\mathbf{S_\tau}, \mathbf{A_\tau}, \theta) = \Pr \left(\mathbf{y_t}=\mathbf{y}) \middle|\mathbf{S_t}=(\mathbf{x_t},\mathbf{\zeta_t},\mathbf{\xi_{t-1}}) \right)
\end{equation}
The log probability of the policy is given by
\begin{equation}
\begin{split}
    \log \pi(\mathbf{S_\tau}, \mathbf{A_\tau}, \theta)  &= \sum_{t_s}\log\lambda^{(\tau)}(t) + \sum_{t_{ns}} \log(1 - \lambda^{(\tau)}(t)) 
\end{split}
\end{equation}
The vector of log firing probabilities $\mathbb{\lambda} = (\lambda(0), ..,\lambda(t))$ can be computed by performing the convolution on the stimulus vector with the filter kernels.

The update equations for each of the policy parameters of the coagent c as per Equation (3) are as follows
\begin{equation}
\begin{split}
\begin{bmatrix}
\mathbf{k} \\
\mathbf{h} \\
\mathbf{l}
\end{bmatrix} 
=  \begin{bmatrix}
\mathbf{k} \\
\mathbf{h} \\
\mathbf{l}
\end{bmatrix}   + \alpha \delta_t \begin{bmatrix}
\nabla_{\mathbf{k}}\log\pi \\
\nabla_{\mathbf{h}}\log\pi \\
\nabla_{\mathbf{l}}\log\pi
\end{bmatrix}
\end{split}
\end{equation}
Where $\delta_t$ is the TD error delivered by the global TD($\lambda$) critic and \begin{equation}
\begin{split}
\begin{bmatrix}
\nabla_{\mathbf{k}}\log\pi \\
\nabla_{\mathbf{h}}\log\pi \\
\nabla_{\mathbf{l}}\log\pi
\end{bmatrix} =\sum_{t \in t_s} \begin{bmatrix}
\mathbf{x_t} \\
\mathbf{\zeta_t} \\
\mathbf{\xi_t}
\end{bmatrix}  (1-\lambda(t)) + \sum_{t \in t_{ns}} \begin{bmatrix}
\mathbf{x_t} \\
\mathbf{\zeta_t} \\
\mathbf{\xi_t}
\end{bmatrix}   (-\lambda(t))
\end{split}
\end{equation}

Equation (9) is also the log likelihood of observing a given spike train response in a spiking coagent. By updates in Equation (10), we are increasing the probability of the spike trains which result in a positive TD error and decreasing the probability of those that result in a negative TD error. Each coagent is thus independently updating its spiking policy in the context of the global error received. In the plain version of the coagent network, each coagent updates as if its spiking policy results in the action selection of the network in iteration of MDP, which is not necessarily the case. This results in high variance in the weight updates. Before we address this issue, we discuss few simplifications and extensions of the above model.

\subsection{Mean-Variance analysis}
In this section, we compare the mean and variance updates of the coagent learning rule with those of backpropagation in an equivalent network. The GLM spiking coagent outputs a probability of spike/no spike for a given time interval. If the spike train time steps are unfolded over time, the computation of the instantaneous firing probabilities can be computed by performing convolution on the state space $\mathbf{s_c}=(\mathbf{x_t},\mathbf{\zeta_t}, \mathbf{\xi_t})$ with the kernels $\mathbf{k}$, $\mathbf{h}$, $\mathbf{l}$ respectively. Instead of sampling from those probabilities, if use them directly for the computation of the responses of the next layer, we essentially have a network similar to that of convolutional neural networks and the convolution operation being differentiable, enables backpropagation through the network. Thus an equivalent network conducive to backpropagation can be construction from the coagent network. The policy of the backpropagation network is
\begin{equation}
   \pi(s,a_k, \mathbf{W}) =  \frac{\exp(o_k)}{\sum_{i=1}^K \exp(o_i)} 
\end{equation}

where a \emph{softmax} action selection is done from the final action probabilities $o_k$ and $\mathbf{W}$ is the parameter matrix of the network.
We now calculate the expected weight update for the kernel $\mathbf{k}_{ij}$, which is the parameter vector relating $i$th input neuron and $j$th hidden neuron. 

\begin{equation}
\begin{split}
    \mathbb{E}[\Delta  \mathbf{k}_{ij}| \mathbf{W}] =& \sum_k \mathbb{E}[\Delta  \mathbf{k}_{ij}| \mathbf{W},a_k] \pi(s,a_k,\mathbf{W})\\
\end{split}
\end{equation}

\begin{equation}
\begin{split}
    \mathbb{E}[\Delta  \mathbf{k}_{ij}| \mathbf{W},a_k] =& \mathbb{E}\left[\alpha \delta_t \frac{\partial \ln \pi(s,a_k,\mathbf{W})}{\partial \mathbf{k}_{ij}}\middle|\mathbf{W}, a=a_k\right]\\
    =& \mathbb{E}\left[\alpha \delta_t | \mathbf{W}, a_k\right] \frac{\partial \ln \pi(s,a_k,\mathbf{W})}{\partial \mathbf{k}_{ij}} 
\end{split}
\end{equation}

If $o_l$ is the $l$th output neuron and $\lambda_j(t)$ is the instantaneous firing probability of neuron $j$ at time $t$, the log derivative of the policy can be written as  

\begin{equation}
\begin{split}
 \frac{\partial \ln \pi(s,a_k)}{\partial \mathbf{k}_{ij}}  =&  \sum_l\frac{\partial \ln \pi(s,a_k)}{\partial o_l}\sum_t\frac{\partial o_l}{\partial \lambda_j(t)}\frac{\partial \lambda_j(t)}{\partial \mathbf{k}_{ij}}  \\
 =&  \sum_t \frac{\partial \lambda_j(t)}{\partial \mathbf{k}_{ij}} \Big(\sum_l  \frac{\partial \ln \pi(s,a_k)}{\partial o_l} \frac{\partial o_l}{\partial \lambda_j(t)}\Big) \\
  =&  \sum_t \lambda_j(t)(1 - \lambda_j(t)) \mathbf{x_i} \Big(\sum_l  \frac{\partial \ln \pi(s,a_k)}{\partial o_l} \frac{\partial o_l}{\partial \lambda_j(t)}\Big)
\end{split}
\end{equation}

\begin{equation}
\begin{split}
    \mathbb{E}[\Delta  \mathbf{k}_{ij}| \mathbf{W}] =& \sum_k \alpha \mathbb{E}\left[\delta_t | \mathbf{W}, a_k\right]  \sum_t \lambda_j(t)(1 - \lambda_j(t)) \mathbf{x_i} \times \Big(\sum_l  \frac{\partial \pi(s,a_k)}{\partial o_l} \frac{\partial o_l}{\partial \lambda_j(t)}\Big)
\end{split}
\end{equation}

The term in the parenthesis in Equation (20) can be seen as the contribution of firing probability of neuron $j$ at time $t$ on the overall action selection.

We will now derive the expected update $\mathbf{k}_{ij}$ in spiking coagent network.
The expected weight update for $\mathbf{k}_{ij}$ given the parameter vector $\mathbf{W}$ is

\begin{equation}
\begin{split}
    \mathbb{E}[\Delta  \mathbf{k}_{ij}| \mathbf{W}] =& \sum_a \sum_{\mathbf{a}_c} \Big(\alpha\mathbb{E}\left[\delta_t \middle|\mathbf{W}, a,\mathbf{a}_c \right]\frac{\partial \ln \pi_c(s,\mathbf{a}_c)}{\partial \mathbf{k}_{ij}} \times \Pr(a | \mathbf{a}_c, \mathbf{W})\Pr(\mathbf{a}_c | \mathbf{W})\Big) \\
    =& \sum_a \alpha\mathbb{E}\left[\delta_t \middle|\mathbf{W}, a \right]\sum_{\mathbf{a}_c} \Big(\frac{\partial \pi_c(s,\mathbf{a}_c)}{\partial \mathbf{k}_{ij}}\Pr(a | \mathbf{a}_c)\Big)
\end{split}
\end{equation}

Equation (21) is simplified using the fact that $ \frac{\partial \ln \pi_c(s,\mathbf{a}_c)}{\partial \mathbf{k}_{ij}} \Pr(\mathbf{a}_c|\mathbf{W} ) = \frac{\partial \ln \pi_c(s,\mathbf{a}_c)}{\partial \mathbf{k}_{ij}} \pi_c(s,\mathbf{a}_c) = \frac{\partial \pi_c(s,\mathbf{a}_c)}{\partial \mathbf{k}_{ij}}$

\begin{equation}
 \frac{\partial \pi_c(s,\mathbf{a}_c)}{\partial \mathbf{k}_{ij}} =  \sum_t y_t \lambda_j(t)(1-\lambda_j(t))\pi_c(s,\mathbf{a}_c-t)\mathbf{x}_i
\end{equation}

where $\pi(s,\mathbf{a}_c-t)$ stands for the policy of the neuron excluding its action at $t$th time step.

\begin{equation}
\begin{split}
&\sum_{\mathbf{a}_c} \Big(\frac{\partial \pi_c(s,\mathbf{a}_c)}{\partial \mathbf{k}_{ij}}\Pr(a | \mathbf{a}_c)\Big)\\
&=\sum_{\mathbf{a}_c}\Pr(a | \mathbf{a}_c) \sum_t \sum_{y_t}y_t\lambda_j(t)(1-\lambda_j(t))\pi(s,\mathbf{a}_c-t)\mathbf{x}_i \\
&= \sum_t \sum_{y_t}y_t\lambda_j(t)(1-\lambda_j(t))\mathbf{x}_i \sum_{\mathbf{a}_c} \pi(s,\mathbf{a}_c-t) \Pr(a|\mathbf{a}_c-t, y_t)\\
&= \sum_t \sum_{y_t}y_t\lambda_j(t)(1-\lambda_j(t))\mathbf{x}_i \mathbb{E}_{\mathbf{a}_c-t} [\Pr(a|y_t)]
\end{split}
\end{equation}

\begin{equation}
\begin{split}
    &\mathbb{E}[\Delta  \mathbf{k}_{ij}| \mathbf{W}] = \sum_a \sum_t \alpha\mathbb{E}\left[\delta_t \middle|\mathbf{W}, a \right] \lambda_j(t)(1-\lambda_j(t)) \mathbf{x}_i
    \times \Big(\mathbb{E}_{\mathbf{a}_c-t}[\Pr(a | y_t=1)]- \mathbb{E}_{\mathbf{a}_c-t}[\Pr(a | y_t=-1)]\Big)
\end{split}
\end{equation}

The last term in parenthesis in Equation (24) is the factor which determines how likely is the firing of the coagent at time $t$, $y_t$,  is to result in action $a$ of the network. Now compare Equation (24) with the corresponding equation (20) from the backpropagation update. The two equations are identical except from the term in the parenthesis. The contribution of the coagent firing at time $t$ is analytically derived in backpropagation whereas in spiking coagents, an identical factor materializes when sampled over multiple trials. Averaged over multiple trials, the expected updates from backpropagation and coagent updates are roughly identical. But the variance of the updates is high in case of coagent networks. From Equation (20), it's clear that for a given parameter vector and state, the only source of stochasticity in backpropagation is from the MDP state transitions. But in case of coagents there are multiple sources of variance for an update (the coagent ($\mathbb{E}_{\mathbf{a}_c}$) and the rest of the network ($\Pr(a|y_t)$ ).

\section {Performance enhancement through variance reduction}
There are multiple techniques to neutralize the sources of variance in a coagent network. One technique is update the learning rule to include an additional factor that is correlated to the contribution of the given coagent to the action selection of the network as in AGREL \cite{agrel}. The learning rule can be modified as 
\begin{equation}
    \Delta \mathbf{k}_{ij} = \Delta \mathbf{k}_{ij} + \alpha \delta_t \frac{\partial \ln \pi(s,a_c,\mathbf{W})}{\partial \mathbf{k}_{ij}} f_o; \quad  f_o \propto \sum_k o_k w_{jk}
\end{equation}
where $f_o$ is the feedback factor to the coagent from the output layer, $o_k$ is the activity of the $k$th coagent in the output layer and $w_{jk}$ is the parameter relating the coagent in the hidden layer with that of the output layer.

In this paper, we chose to avoid the feedback factors which might include the information about the weights in the rest of the network but instead purely focus on locally updating the coagents with minimal feedback from rest of the network. We focus on the variance reduction by solely targeting architectural design.

\subsection{Variance reduction through a modular connectionist architecture}

Brain networks have been demonstrated to have the property of hierarchical modularity, i.e, each module being composed of sub-modules which are in turn composed of several sub-modules. This modular structure is claimed to be responsible for faster adaptation and evolution of the system with changing stimulus conditions (\cite{modular}). \cite{jacobs} showed that incorporation of a modular architecture in neural networks results in faster learning compared to a fully connected architecture by decomposing the task into many functionally independent tasks. In this study we demonstrate that such a modular architecture is conducive to local learning rules.

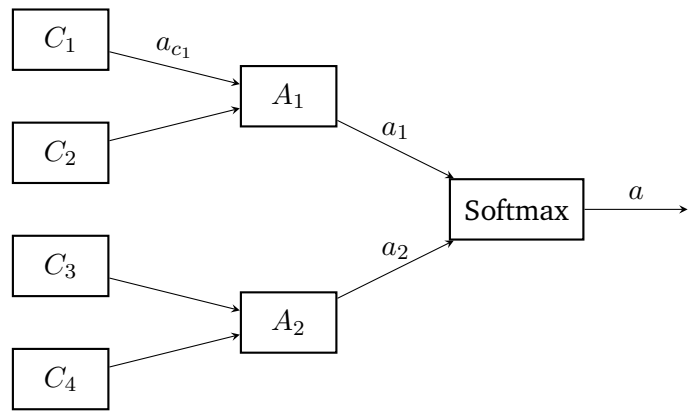
\begin{figure}[htpb]
\centering

\begin{tikzpicture}[scale=1.50]
  \node (rect) (1) at (0,2) [draw,thick,minimum width=1cm,minimum height=0.8cm,text width=1cm,align=center] {$C_1$};
  \node (rect) (2) at (0,1) [draw,thick,minimum width=1cm,minimum height=0.8cm,text width=1cm,align=center] {$C_2$};
  \node (rect) (3) at (0,0) [draw,thick,minimum width=1cm,minimum height=0.8cm,text width=1cm,align=center] {$C_3$};
  \node (rect) (4) at (0,-1) [draw,thick,minimum width=1cm,minimum height=0.8cm,text width=1cm,align=center] {$C_4$};
  \node (rect) (5) at (2,1.5) [draw,thick,minimum width=1cm,minimum height=0.8cm,text width=1cm,align=center] {$A_1$};
  \node (rect) (6) at (2,-0.5) [draw,thick,minimum width=1cm,minimum height=0.8cm,text width=1cm,align=center] {$A_2$};
  \node (rect) (7) at (4,0.5) [draw,thick,minimum width=0.5 cm,minimum height=0.8cm,text width=1.5cm,align=center] {Softmax};
  \path [<-,>=stealth](5) edge node[above]{$a_{c_1}$} (1);
  \path [<-,>=stealth](5) edge (2);
  \path [<-,>=stealth](6) edge (3);
  \path [<-,>=stealth](6) edge (4);
  \path [<-,>=stealth](7) edge node[above]{$a_1$} (5);
  \path [<-,>=stealth](7) edge node[above]{$a_2$}(6);
  \path [->,>=stealth](7) edge node[above]{$a$}(5.5,0.5);

\end{tikzpicture}

\caption{Modular connectionist architecture reduces variance by attributing each coagent to one action thereby increasing the probability $\Pr(a|a_{c_1})$ over a fully connected architecture  }
\end{figure}

Figure 2 shows a modular connectionist architecture with sparse modular connections instead of a fully-connected architecture. In the figure it can be seen that any coagent in the hidden layer is connected to only one coagent in the output layer. This can be extended to multiple hidden layers by decomposing the coagents in a layer into modules where a given module is only connected to the corresponding module in the succeeding layer. \cite{plaut} showed that learning in such a modular architecture is achieved much faster than in a fully connected architecture. 

In case of a modular network, when a TD error is observed upon the selection of an action, the coagents in all the layers belonging to the action module are updated. In order to increase the speed of learning, we chose to appropriately update the coagents belonging to the other action modules as well. If an action that is the desired action is fired, all the coagents responsible for the firing of that action are penalized with a negative TD error. On the other hand if the action is not fired, the coagents corresponding to that action module are rewarded with a positive TD error. This is to ensure that for any given state only the desired action is fired and the rest of actions are silent. Thus all the coagents are updated during any iteration of an MDP.

\subsection{Variance reduction through population coding}
As we discussed earlier, on averaging over multiple trials the weight updates in the coagent network approximate the contribution of the weights to the overall action selection. An obvious variance reduction technique would be to average the weight updates over multiple trials. This averaging can be done either by running the same network in multiple trials or run multiple networks in parallel and select the action based on the ensemble activity. This form of encoding of actions from the joint activity of a population of neurons is termed as population coding. Experimental evidence supports that this coding technique is widely employed in sensor and motor areas of the brain \cite{population}.

We run a population of networks to get the spike responses from the output neurons which are then averaged across the networks to give the final output probabilities. The action is chosen from the output vectors by applying softmax function. While performing updates on the individual networks, the TD error is delivered to the network  as it is, if the action chosen by the network is same as the final action of the ensemble, else the TD error is delivered with its sign reversed. These updates are off-policy as the action executed by the ensemble might not be the action chosen by the current network's policy.

\section{Reparameterization trick in spiking neural networks}

In the previous chapter we introduced a local learning algorithm for training spiking neural networks. This algorithm, however, is susceptible high degree of variance which we attempted to mitigate through architectural variations. In this chapter, we introduce a second technique of training spiking neural networks which does not solely rely on local information but instead is more closely related to backpropagation.

As we discussed before, applying a backpropagation algorithm to spiking neural networks is not feasible owing to the discrete nature of its information transmission. To overcome this hurdle, we employ the technique developed in variational inference to facilitate backpropagation through a stochastic node: the reparameterization trick (\cite{kingma}). We model the policy of a spiking neuron as a probability distribution which generates spikes through sampling. We then apply the reparameterization trick to backpropagate through the samples to assign credit/blame across individual neurons.

The reparameterization trick enables us to model the randomness in sampling as an input to the model rather than attributing it to the model parameters thus rendering all the model parameters continuous and differentiable and thereby facilitating backpropagation. 

\subsection{Related work}
In this section, we review approaches in literature that attempts to apply backpropagation to train spiking neural networks. \cite{pfeiffer} considered membrane voltage potentials of spiking neurons as differentiable signals where discontinuities at spike times are considered as noise. This enables backpropagation that works directly on spike signals and membrane potentials. \cite{huh} formulated a differentiable synapse model of a spiking neuron and derived an exact gradient calculation. \cite{bohte} introduced the algorithm SpikeProp with the target of learning a set of firing times at output neurons given the input patterns. The algorithm backpropagates on the error function of aggregate difference between desired spike times and actual spike times. Similarly \cite{mostafa} uses a temporal coding scheme where information is encoded in spike times instead of spike rates, the network input-output relation is differentiable almost everywhere. In \cite{kherad}, the network uses a form of temporal coding called rank-order coding. In this coding technique, a spiking neuron is limited to one spike per neuron but the firing order among the neurons carries relevant information. In this paper, an algorithm akin to backpropagation called S4NN is derived. 

To our best knowledge, this is the first work of literature that applies the reparameterization trick to backpropagate errors through spiking neural networks.

\subsection{Reparameterization trick}

Consider the following expectation, where a discrete random variable $z$ is sampled from a distribution $p_\theta(z)$ which depends on $\theta$ and $f_\theta(z)$ is a cost function.
\begin{equation}
  \mathbb{E}_{z \sim p_\theta(z)}[f_\theta(z)]  
\end{equation}

In order to find the best parameter $\theta$ to minimize the above expectation, we need to compute its derivative 
\begin{equation}
    \nabla_\theta \mathbb{E}_{z \sim p_\theta(z)}[f_\theta(z)]
\end{equation}

To make the above derivative differentiable with respect to $\theta$, the random variable $z$ is expressed as a differentiable function of deterministic variable with an additive noise as given below
\begin{flalign}
z &= g_\phi(x,\epsilon) \\
\epsilon &\sim p^{'}(\epsilon)
\end{flalign}
where $\epsilon$ is an additive random variable which is sampled from a probability distribution $p^{'}(\epsilon)$. Here $x$ is the parameter of the model which is deterministic and hence is differentiable and $\epsilon$ is a noise term that now accounts for the randomness of the model. The derivative in (3.2) can now be computed as follows
\begin{flalign}
\nabla_\theta \mathbb{E}_{z \sim p_\theta(z)}[f_\theta(z)] &= \nabla_\theta \mathbb{E}_{\epsilon \sim p^{'}(\epsilon)}[f_\theta(g_\phi(x,\epsilon))] \\
&= \mathbb{E}_{\epsilon \sim p^{'}(\epsilon)}[\nabla_\theta f_\theta(g_\phi(x,\epsilon))] \\
&\approx \frac{1}{L}\sum_{l=1}^L \nabla_\theta f_\theta(g_\phi(x,\epsilon^{(l)}))
\end{flalign}

The choice of $g_\phi$ can be any convenient distribution such as normal distribution but in our case we choose a special function called gumbel softmax.

\subsection{Reparameterization through gumbel-softmax}
Actions for a spiking neuron are sampled from a categorical distribution with probabilities associated with each action category. Sampling from a categorical distribution is typically a non-differentiable function. In \cite{gumbel} a Gumbel-softmax function is introduced which provides a way to extract differentiable samples from a categorical distribution.

Gumbel-softmax function generates samples as follows:
\begin{equation}
    y_i = \frac{\exp((\log(\pi_i) + g_i)/\tau)}{\sum_{j=1}^k \exp((\log(\pi_j) + g_j)/\tau)}
\end{equation}
where $g_i$ is an iid sample from Gumbel(0,1) distribution which is generated by sampling $u$ from a normal distribution $\mathcal N(0,1)$ and calculating $g(u) = -\log(-\log(u))$. The above equation generates categorical samples $y_i$ from log probabilities of actions $\pi_i$ of a policy.

\subsection{Spiking neuron as a stochastic node}
To apply the reparameterization trick, spiking neuron needs to be modeled as a stochastic node with actions sampled from a probability distribution. Any of the spiking neuron models described in the previous chapter can be used to model the neuron. As a proof-of-concept, we use the simple version of a memoryless spiking neuron.

The policy of the neuron is given as 
\begin{equation}
  \pi(s,a=a_k) = \frac{\exp(a_k(b_i + \sum_j W_{ij} \sigma_j))}{\sum_{-2}^2\exp(a_i(b_i + \sum_j W_{ij} \sigma_j))}
\end{equation}

where $b_i$, $W_{ij}$ are the parameters of the policy.
The outputs of a layer are then sampled from the log probabilities of the policies using gumbel-softmax function as shown below
\begin{equation}
    y_i = \frac{\exp((\log(\pi(s,a_k)) + g_i)/\tau)}{\sum_{j=1}^k \exp((\log(\pi(s,a_j) + g_j)/\tau)}
\end{equation}

Assuming the next layer transforms the outputs $y_i$ as $f(y_i)$. The differential with respect to a policy parameter $W_{ij}$ can be computed as follows:
\begin{equation}
   \frac{\partial f(y_i)}{\partial W_{ij}} =  \frac{\partial f(y_i)}{\partial y_i} \frac{\partial y_i}{\partial \pi(s,a)}\frac{\partial \pi(s,a)}{\partial W_{ij}}
\end{equation}

\subsection{Network implementation}
The actor and the critic networks are both multi-layered neural network which shared initial layers. An advantage actor-critic learning technique \cite{a3c} is used to optimize the network where the policy gradient updates are made using the advantage function.

\section{Case Studies}

In this section, we train spiking neural networks using the techniques discussed in the previous sections and apply them in various contexts of reinforcement learning. We compare and contrast between the local learning framework developed in this study and backpropagtion with reparameterization technique.

\subsection{Reinforcement learning domains}
The following RL domains ((covering delayed rewards and continuous control settings) are used in this study.

\subsubsection{Gridworld 5$\times$5}
In this domain the agent has to navigate a maze-like environment to reach a terminal state by learning the path that provides the maximum possible reward. The agent has four options for mobility (UP, DOWN, LEFT, RIGHT). The agent moves in the specified direction with a probability of 0..8.  With probability 0.05 the agent veers to the right from its intended direction and veers to the left with a probability of 0.05. The agent does not execute an action with the probability of 0.1. If the agent attempts to move in a direction that puts it beyond the boundaries of the domain or hits an obstacle the agent remains stationary. The agent starts in state 1, and the process ends when the agent reaches state 23.
\begin{figure}[htpb]
\centering
\includegraphics[width=80mm]{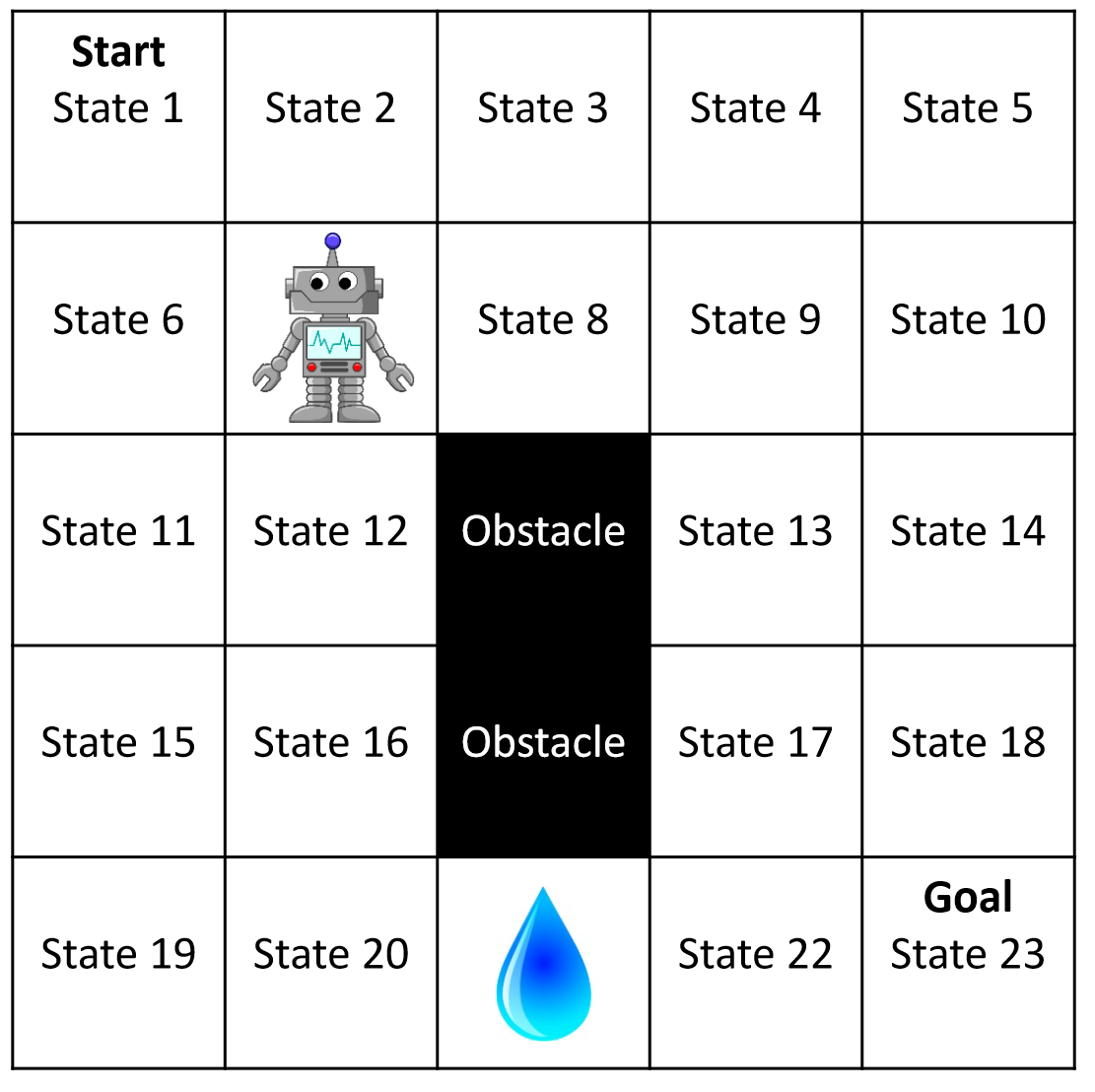}
\caption{Gridworld derived from \cite{notes}}
\end{figure}  

\subsubsection{Gridworld 10$\times$10}
The $10 \times 10$ gridworld domain is a scaled up version of the $5\times 5$ domain but without the obstacles. The state at $(10,10)$ is the terminal state with a reward of 10. Every other transition has a reward of zero. The environment has stochastic actions where a given action is executed with a probability of $0.8$, the agent veers left with a probability $0.05$, veers right with a probability $0.05$ and stays in the same position with the probability $0.1$.
\subsubsection{Cartpole}
The Cart-pole environments consists of two interacting bodies:  a cart with position $x$ and velocity $v$, and a pole with angle $\theta$ and angular velocity $\omega$. The state vector consists of these continuous variables $(x,v,\theta, \omega)$ with dynamics described in \cite{cartpole}. The task is to balance the pole for $200$ time steps with two possible actions and reward of $1$ for each time step that the pole remains balanced.
\subsubsection{Mountain Car}
In this domain, the task for the agent is to get a car is stuck in a valley to the top of the hill in front of the car. The agent has three possible actions Forward, Reverse, Neutral. The reward is -1 for every time step till the car reaches the top of the hill.

\subsection{Experiments}

\subsubsection{PGCN learning in gridworld with a memoryless spiking neuron}

This study is performed on $5 \times 5$ gridworld task. The 23 states are represented using 7 input neurons in binary coding, with +1 for firing and -1 for silent. The hidden layer consists of 10 neurons and the output layer has 4 neurons each representing a different action from the action space. For reduction in variance 10 such networks are run in parallel and their outputs are averaged to give an average firing rate for each of the action. A softmax function is then applied to the output firing rates to chose an action. The learning curve thus obtained is contrasted with the curve obtained through backpropagation using a similar architecture as shown in the figure. 

\begin{figure}[htpb]
\centering
\includegraphics[width=120mm]{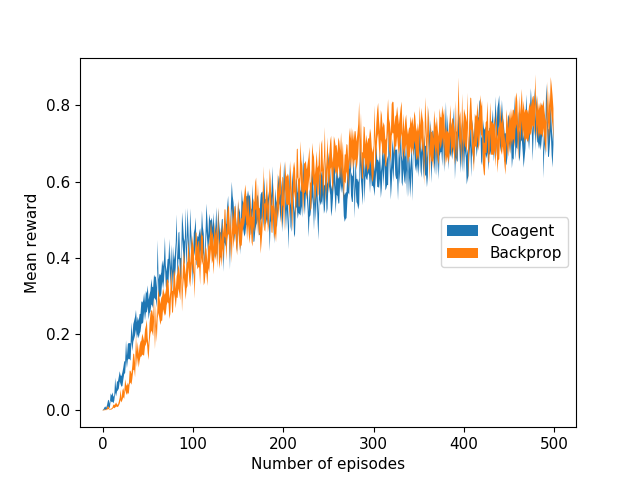}
\caption{Case study: Gridworld $5 \times 5$. Learning curve of the spiking coagent network against the baseline of backpropagation}
\end{figure}

\subsubsection{PGCN learning in gridworld with a GLM spiking neuron}

This study is performed on $10 \times 10$ gridworld task. The $100$ states of gridworld are encoded using $3$ neurons with a spike train length of $5$. The hidden layer has $5$ coagents each with a spike train length of $3$. The stimulus filter $\mathbf{k}$ of a coagent is a kernel of $3$ parameters which produces the hidden layer spike train responses upon convolution with the spike train stimuli from the previous layer. For simplicity we ignore the other filters. The output layer has $4$ coagents each corresponding to an action of the domain and the activity of the coagent is encoded in a single spike. The policy of each of the coagents is as described in Equation (10). The coagents are organized in a modular connectionist architecture described in the previous section. A population of $10$ such networks are concurrently used to select the actions. 

An advantage actor-critic network is used as a baseline for this domain. The state is encoded in the similar manner as the spiking coagent with the temporal component flattened to a spatial component. The actor network is a three layered network with the input layer consists of 15 neurons, hidden layer of 100 neurons and an output layer of 4 neurons. Critic is also a neural network with the same architecture as that of the actor network but with the output layer has one node representing the value function.
We also compare both the spiking agent actor-critic and advantage actor-critic with a tabular actor-critic as shown in Figure. From the figure, we can see that the tabular actor-critic works well for the simple task of gridworld but spiking agent actor-critic is close in performance to advantage actor-critic.

\begin{figure}[htpb]
         \centering
	 \includegraphics[width=120mm]{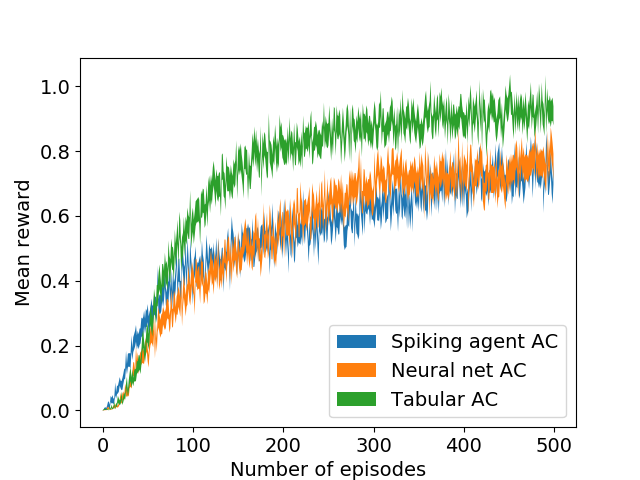}
         \caption{Performance  in  the  gridworld task with a modular spiking agent actor-critic (AC) framework (10 networks) against the baselines of advantage actor-critic and tabular AC}
     \end{figure}

\subsubsection{PGCN learning in mountain car}
We apply the memoryless spiking neuron model to the mountain car task. The actor network is a three-layered neural network with 20 neurons in the input layer, 50 neurons in the hidden layer and three neurons in the output layer. In the input layer, 10 neurons represent the state and the remaining 10 neurons represent the velocity. The continuous state variables are represented in binary-coded input layer. The output layer accounts for the three possible actions from the action space. The actions are selection by averaging averaging across 10 such networks. Figure shows the learning achieved in the mountain car task.

\begin{figure}[htpb]
         \centering
	 \includegraphics[width=120mm]{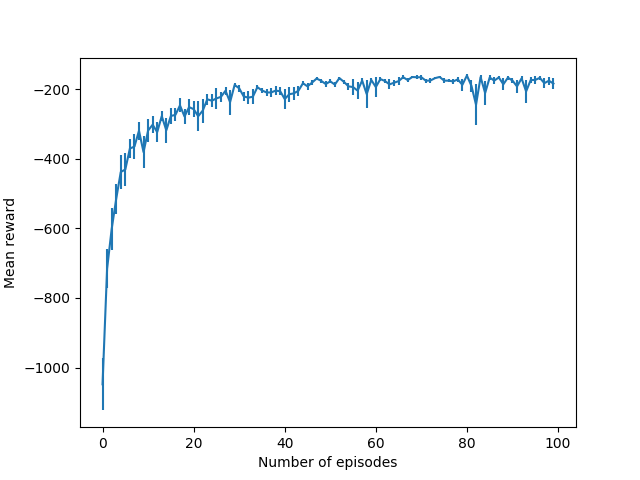}
         \caption{Performance of the PGCN update rule in the mountain car task}
     \end{figure}

\subsubsection{PGCN learning in cart pole: Comparison with Hebbian learning rule}
We now apply the memoryless spiking neuron model to the cart pole task. We use $4$ input neurons to represent the value of each of the $4$ state variables. The input neurons are not of spiking nature but instead represent the variables in continuous form. The hidden layer has $200$ spiking agents and the output layer has $2$ agents representing the two actions in the action space. As before, actions are chosen using a population of $10$ such networks.

As we saw in Equation(15), the weight updates for the single spike model are equivalent to Hebbian updates factored by gradients. We now determine the role the gradient factor plays in the convergence to an optimal policy. Here we compare the performance of the coagent policy gradients with that of local Hebbian updates in a network with binary stochastic spikes. Hebbian rule updates the weight of the synapse according to the below equation
\begin{equation}
    \Delta w_{ij} =  \alpha x_i x_j
\end{equation}
where $x_i$, $x_j$ are the spiking activities of the pre and post-synaptic neuron. 

 We use $4$ input neurons to represent the value of each of the $4$ state variables. The hidden layer has $200$ coagents and the output layer has $2$ coagents for the two actions. Actions are chosen using a population of $10$ such networks.

Figure (5a) shows the comparison of coagent updates with that of simple Hebbian correlation updates. Hyperparameters of learning rate schedule and momentum are tuned separately for each experiment. It can be seen that Hebbian updates result in a high variance and that local policy gradients are pivotal to the convergence to an optimal policy.

\begin{figure}[htpb]
         \centering
	 \includegraphics[width=120mm]{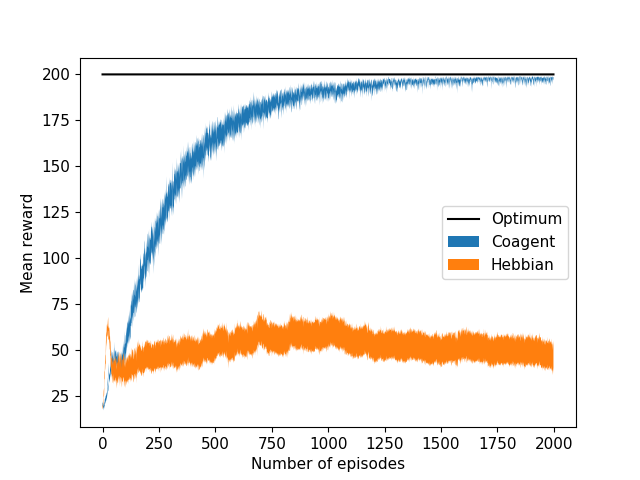}
         \caption{Comparison of coagent updates against local Hebbian updates on cartpole task.}
     \end{figure}

\subsubsection{Effect of modular architecture on learning performance}
In this experiment, we test the performance enhancement in the cartpole achieved by employing a modular architecture. As a proof-of-concept, we train the coagent network first with a fully connected network and then with the modular connectionist architecture shown in the Figure on the same task. All the curves are obtained by averaging over 10 networks. Figure (5b) shows the comparison of the learning curves from the two architectures. It can be seen that a fully connected network barely learns the policy whereas the modular architectures converges to an optimal policy. 
\begin{figure}[htpb]
         \centering
	 \includegraphics[width=120mm]{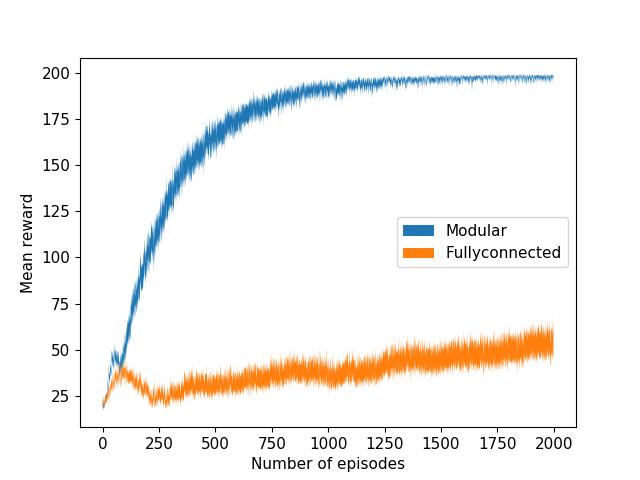}
         \caption{Performance improvement of modular connectionist architecture from Figure 3 against a fully connected architecture in the cartpole task.}
     \end{figure}

\subsubsection{Effect of population coding on learning performance}

In this experiment, we demonstrate the effectiveness of population coding as a variance reduction technique. We apply the coagent network to the cartpole task by averaging over varying number of networks. We first obtain a learning curve by using just one network and then gradually increase the population size to demonstrate how increasing the population size reduces the variance in learning and improves its performance in solving the task. We use a modular architecture for all the settings. Figure shows the improvement in performance achieved by increasing the population size in steps of 1, 5, 10 and 20.
     
\begin{figure}[htpb]
         \centering
	 \includegraphics[width=120mm]{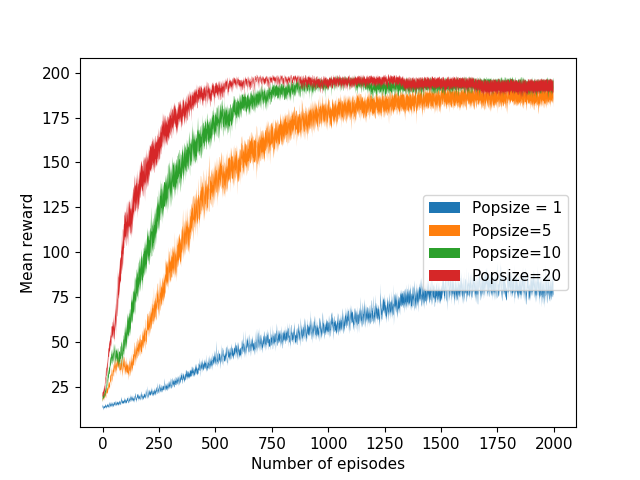}
         \caption{Effect of population coding on the performance in the cartpole task. Increase in population size increases the speed of convergence to the optimum}
     \end{figure}
     
 \subsubsection{Code}

The code for the above experiments can be found at \href{https://github.com/asneha213/spiking-agent-RL}{https://github.com/asneha213/spiking-agent-RL}

\section{Discussion \& Conclusion}
In this study, we introduced two techniques for training spiking neural networks to perform reinforcement learning tasks. In the first technique, we extended the concept of a hedonistic neuron  \cite{hedonistic}, \cite{seung} by formulating a spiking neuron as an RL agent. We explored various models of a spiking neuron to efficiently model the neuron as an RL agent. The generalized linear model while being computationally tractable can closely model most of the features of a biological neuron. We also used a memoryless spiking neuron model as a proof-of-concept to validate some of our claims. Although a powerful representational technique, our learning framework suffers from high variance in convergence, which is inevitable in all local learning paradigms. To mitigate this issue, we analyzed the mean and variance of our updates against the backpropagatation updates from an equivalent network. From this analysis, we updated our framework by employing variance reduction techniques to ensure competitive performance compared to traditional optimization techniques. In a different technique, we modeled the spiking neuron as a stochastic node with actions sampled from a probability distribution and used reparameterization trick from variational inference to backpropagate through the spiking neural network.

In this study, we worked with feed forward networks without any recurrent connections. \cite{kostas} extended the PGCN framework to account for asynchronous updates in case of recurrent connections. Extending this work to incorporate recurrent inhibitory connections is a possible future direction. \cite{noisy} theoretically proved that noisy spiking neuron with temporal coding has more computational power than sigmoidal neuron. Developing architectures and learning rules to harness that computational power can possibly have implications in the advancement of artificial intelligence as well as shed light on the functioning of the human brain.





\bibliography{literature.bib} 

@article {gasper,
	author = {Tka{\v c}ik, Ga{\v s}per and Prentice, Jason S. and Balasubramanian, Vijay and Schneidman, Elad},
	title = {Optimal population coding by noisy spiking neurons},
	volume = {107},
	number = {32},
	pages = {14419--14424},
	year = {2010},
	doi = {10.1073/pnas.1004906107},
	publisher = {National Academy of Sciences},
	issn = {0027-8424},
	URL = {https://www.pnas.org/content/107/32/14419},
	eprint = {https://www.pnas.org/content/107/32/14419.full.pdf},
	journal = {Proceedings of the National Academy of Sciences}
}

@article{modular,
	title = {Modular and {Hierarchically} {Modular} {Organization} of {Brain} {Networks}},
	volume = {4},
	language = {en},
	journal = {Frontiers in Neuroscience},
	author = {Meunier, David and Lambiotte, Renaud and Bullmore, Edward T.},
	year = {2010},
}

@misc{spikingneuron,
    title={Spiking Neuron Models},
    author={Wulfram Gerstner and Werner M. Kistler},
    year={2002}
}

@misc{notes,
    title={CMPSCI 687: Reinforcement Learning Fall 2018 Class Syllabus, Notes, and Assignments},
    author={Philip S. Thomas},
    year={2018}
}

@misc{kingma,
    title={Auto-Encoding Variational Bayes},
    author={Diederik P Kingma and Max Welling},
    year={2013},
    eprint={1312.6114},
    archivePrefix={arXiv},
    primaryClass={stat.ML}
}

@article{gumbel,
  title={Categorical Reparameterization with Gumbel-Softmax},
  author={Eric Jang and Shixiang Gu and Ben Poole},
  journal={ArXiv},
  year={2017},
  volume={abs/1611.01144}
}

@misc{mostafa,
    title={Supervised learning based on temporal coding in spiking neural networks},
    author={Hesham Mostafa},
    year={2016},
    eprint={1606.08165},
    archivePrefix={arXiv},
    primaryClass={cs.NE}
}

@misc{kherad,
    title={S4NN: temporal backpropagation for spiking neural networks with one spike per neuron},
    author={Saeed Reza Kheradpisheh and Timothée Masquelier},
    year={2019},
    eprint={1910.09495},
    archivePrefix={arXiv},
    primaryClass={cs.NE}
}

@article{bohte,
	title = {Error-backpropagation in temporally encoded networks of spiking neurons},
	volume = {48},
	issn = {0925-2312},
	url = {http://www.sciencedirect.com/science/article/pii/S0925231201006580},
	doi = {https://doi.org/10.1016/S0925-2312(01)00658-0},
	number = {1},
	journal = {Neurocomputing},
	author = {Bohte, Sander M. and Kok, Joost N. and Poutré, Han [La},
	year = {2002},
	pages = {17 -- 37}
}

@misc{huh,
    title={Gradient Descent for Spiking Neural Networks},
    author={Dongsung Huh and Terrence J. Sejnowski},
    year={2017},
    eprint={1706.04698},
    archivePrefix={arXiv},
    primaryClass={q-bio.NC}
}

@ARTICLE{pfeiffer,
  
AUTHOR={Lee, Jun Haeng and Delbruck, Tobi and Pfeiffer, Michael},   
	 
TITLE={Training Deep Spiking Neural Networks Using Backpropagation},      
	
JOURNAL={Frontiers in Neuroscience},      
	
VOLUME={10},      

PAGES={508},     
	
YEAR={2016},      
	  
URL={https://www.frontiersin.org/article/10.3389/fnins.2016.00508},       
	
DOI={10.3389/fnins.2016.00508},      
	
ISSN={1662-453X}
}

@article{agrel,
	title = {Attention-{Gated} {Reinforcement} {Learning} of {Internal} {Representations} for {Classification}},
	volume = {17},
	issn = {0899-7667, 1530-888X},
	url = {http://www.mitpressjournals.org/doi/10.1162/0899766054615699},
	doi = {10.1162/0899766054615699},
	language = {en},
	number = {10},
	urldate = {2019-06-18},
	journal = {Neural Computation},
	author = {Roelfsema, Pieter R. and Ooyen, Arjen van},
	month = oct,
	year = {2005},
	pages = {2176--2214}
}

@article{kalman,
	title = {All learning is {Local}: {Multi}-agent {Learning} in {Global} {Reward} {Games}},
	language = {en},
	author = {Chang, Yu-han and Ho, Tracey and Kaelbling, Leslie P},
	pages = {8}
}

@article{glm,
	title = {Learning {First}-to-{Spike} {Policies} for {Neuromorphic} {Control} {Using} {Policy} {Gradients}},
	url = {http://arxiv.org/abs/1810.09977},
	language = {en},
	urldate = {2019-07-15},
	journal = {arXiv:1810.09977 [cs, eess, stat]},
	author = {Rosenfeld, Bleema and Simeone, Osvaldo and Rajendran, Bipin},
	month = oct,
	year = {2018},
	note = {arXiv: 1810.09977}
}

@article{pillow,
	title = {Spatio-temporal correlations and visual signalling in a complete neuronal population},
	volume = {454},
	url = {https://doi.org/10.1038/nature07140},
	journal = {Nature},
	author = {Pillow, Jonathan W. and Shlens, Jonathon and Paninski, Liam and Sher, Alexander and Litke, Alan M. and Chichilnisky, E. J. and Simoncelli, Eero P.},
	month = jul,
	year = {2008},
	pages = {995}
}

@article{truccolo,
	title = {A {Point} {Process} {Framework} for {Relating} {Neural} {Spiking} {Activity} to {Spiking} {History}, {Neural} {Ensemble}, and {Extrinsic} {Covariate} {Effects}},
	volume = {93},
	issn = {0022-3077, 1522-1598},
	url = {http://www.physiology.org/doi/10.1152/jn.00697.2004},
	doi = {10.1152/jn.00697.2004},
	language = {en},
	number = {2},
	urldate = {2019-08-03},
	journal = {Journal of Neurophysiology},
	author = {Truccolo, Wilson and Eden, Uri T. and Fellows, Matthew R. and Donoghue, John P. and Brown, Emery N.},
	month = feb,
	year = {2005},
	pages = {1074--1089}
}

@inproceedings{comdp,
  author    = {Philip S. Thomas and
               Andrew G. Barto},
  title     = {Conjugate Markov Decision Processes},
  booktitle = {Proceedings of the 28th International Conference on Machine Learning,
               {ICML} 2011, Bellevue, Washington, USA, June 28 - July 2, 2011},
  pages     = {137--144},
  year      = {2011},
  bibsource = {dblp computer science bibliography, https://dblp.org}
}

@article{williams,
  author = {Williams, R. J.},
  description = {CCNLab BibTeX},
  journal = {Machine Learning},
  pages = {229-256},
  title = {Simple statistical gradient-following algorithms for connectionist
	reinforcement learning},
  volume = 8,
  year = 1992
}

@article{jacobs,
  author =	"R. A. Jacobs and M. I. Jordan and A. G. Barto",
  title =	"Task Decomposition Through Competition in a Modular
		 Connectionist Architecture: The What and Where Vision
		 Task",
  journal =	"Cognitive Science",
  year = 	"1991",
  volume =	"15",
  pages =	"219--250",
}

@article{cartpole,
	title = {Correct equations for the dynamics of the cart-pole system},
	language = {en},
	author = {Florian, Razvan V},
	pages = {6}
}

@Article{plaut,
  author =	"D.C. Plaut and G.E. Hinton",
  title =	"Learning Sets of Filters Using Back-Propagation",
  journal =	"Computer Speech and Language",
  year = 	"1987",
  volume =	"2",
  pages =	"35--61",
  ref =  	"J19",
}

@article {optimal,
	author = {Tka{\v c}ik, Ga{\v s}per and Prentice, Jason S. and Balasubramanian, Vijay and Schneidman, Elad},
	title = {Optimal population coding by noisy spiking neurons},
	volume = {107},
	number = {32},
	pages = {14419--14424},
	year = {2010},
	doi = {10.1073/pnas.1004906107},
	publisher = {National Academy of Sciences},
	issn = {0027-8424},
	URL = {https://www.pnas.org/content/107/32/14419},
	eprint = {https://www.pnas.org/content/107/32/14419.full.pdf},
	journal = {Proceedings of the National Academy of Sciences}
}

@Article{asn,
  author =	"A. G. Barto and R. S. Sutton and P. S. Brouwer",
  year = 	"1981",
  title =	"Associative search network: a reinforcement learning
		 associative memory",
  journal =	"Biological Cybernetics",
  volume =	"40",
  number =	"3",
  pages =	"201--211"
}

@Unpublished{qlearning,
  author =	"C. Watkins and P. Dayan",
  title =	"{Q}-Learning",
  note = 	"to appear in the Journal of Machine Learning",
  year = 	"1992",
  ref =  	"PP32",
}

@InProceedings{pgcn,
  title =	"Policy Gradient Coagent Networks",
  author =	"Philip S. Thomas",
  bibdate =	"2014-12-10",
  bibsource =	"DBLP,
		 http://dblp.uni-trier.de/db/conf/nips/nips2011.html#Thomas11",
  booktitle =	"Advances in Neural Information Processing Systems 24:
		 25th Annual Conference on Neural Information Processing
		 Systems 2011. Proceedings of a meeting held 12-14
		 December 2011, Granada, Spain",
  year = 	"2011",
  booktitle =	"NIPS",
  editor =	"John Shawe-Taylor and Richard S. Zemel and Peter L.
		 Bartlett and Fernando C. N. Pereira and Kilian Q.
		 Weinberger",
  pages =	"1944--1952",
  URL =  	"http://papers.nips.cc/book/advances-in-neural-information-processing-systems-24-2011",
}

@Misc{xie,
  title =	"Learning in neural networks by reinforcement of
		 irregular spiking",
  author =	"Xiaohui Xie and H. Sebastian Seung",
  year = 	"2004",
  bibsource =	"OAI-PMH server at citeseerx.ist.psu.edu",
  contributor =  "CiteSeerX",
  description =  "spiking",
  language =	"en",
  URL = "http://citeseerx.ist.psu.edu/viewdoc/summary?doi=10.1.1.72.4972;http://hebb.mit.edu/people/seung/papers/sp9.pdf",
}

@Article{deeprl,
  title =	"Human-level control through deep reinforcement
		 learning",
  author =	"Volodymyr Mnih and Koray Kavukcuoglu and David Silver
		 and Andrei A. Rusu and Joel Veness and Marc G.
		 Bellemare and Alex Graves and Martin A. Riedmiller and
		 Andreas Fidjeland and Georg Ostrovski and Stig Petersen
		 and Charles Beattie and Amir Sadik and Ioannis
		 Antonoglou and Helen King and Dharshan Kumaran and Daan
		 Wierstra and Shane Legg and Demis Hassabis",
  journal =	"Nature",
  year = 	"2015",
  number =	"7540",
  volume =	"518",
  bibdate =	"2018-11-14",
  bibsource =	"DBLP,
		 http://dblp.uni-trier.de/https://doi.org/10.1038/nature14236;
		 DBLP,
		 http://dblp.uni-trier.de/https://www.wikidata.org/entity/Q27907579;
		 DBLP,
		 http://dblp.uni-trier.de/db/journals/nature/nature518.html#MnihKSRVBGRFOPB15",
  pages =	"529--533",
}

@article{dopamine,
  author = {Schultz, W. and Dayan, P. and Montague, P. R.},
  description = {CCNLab BibTeX},
  journal = {Science},
  pages = 1593,
  title = {A Neural Substrate of Prediction and Reward.},
  volume = 275,
  year = 1997
}

@Article{kostas,
  title =	"Asynchronous Coagent Networks: Stochastic Networks for
		 Reinforcement Learning without Backpropagation or a
		 Clock",
  author =	"James Kostas and Chris Nota and Philip S. Thomas",
  journal =	"CoRR",
  year = 	"2019",
  volume =	"abs/1902.05650",
  bibdate =	"2019-03-02",
  bibsource =	"DBLP,
		 http://dblp.uni-trier.de/db/journals/corr/corr1902.html#abs-1902-05650",
  URL =  	"http://arxiv.org/abs/1902.05650",
}

@Misc{noisy,
  title =	"Noisy Spiking Neurons with Temporal Coding have more
		 Computational Power than Sigmoidal Neurons",
  author =	"Wolfgang Maass",
  year = 	"1997",
  month =	jul # "~04",
  bibsource =	"OAI-PMH server at cs1.ist.psu.edu",
  language =	"en",
  oai =  	"oai:CiteSeerPSU:180285",
  rights =	"unrestricted",
  URL =  	"http://citeseer.ist.psu.edu/180285.html;
		 http://www.cis.tu-graz.ac.at/igi/maass/i5.ps.gz",
}

@article{glmfig,
	title = {Capturing the dynamical repertoire of single neurons with generalized linear models},
	url = {http://arxiv.org/abs/1602.07389},
	language = {en},
	urldate = {2019-08-12},
	journal = {arXiv:1602.07389 [q-bio]},
	author = {Weber, Alison I. and Pillow, Jonathan W.},
	month = feb,
	year = {2016},
	note = {arXiv: 1602.07389}
}

@article{population,
	title = {Functional properties of neurons in middle temporal visual area of the macaque monkey. {I}. {Selectivity} for stimulus direction, speed, and orientation},
	volume = {49},
	issn = {0022-3077, 1522-1598},
	url = {http://www.physiology.org/doi/10.1152/jn.1983.49.5.1127},
	doi = {10.1152/jn.1983.49.5.1127},
	language = {en},
	number = {5},
	urldate = {2019-08-10},
	journal = {Journal of Neurophysiology},
	author = {Maunsell, J. H. and Van Essen, D. C.},
	month = may,
	year = {1983},
	pages = {1127--1147}
}

@Misc{a3c,
  title =	"Asynchronous Methods for Deep Reinforcement Learning",
  author =	"Volodymyr Mnih and Adri{\`a} Puigdom{\`e}nech Badia
		 and Mehdi Mirza and Alex Graves and Timothy P.
		 Lillicrap and Tim Harley and David Silver and Koray
		 Kavukcuoglu",
  year = 	"2016",
  month =	jun # "~16",
  bibsource =	"OAI-PMH server at export.arxiv.org",
  identifier =	"ICML 2016",
  oai =  	"oai:arXiv.org:1602.01783",
  subject =	"Computer Science - Learning",
  URL =  	"http://arxiv.org/abs/1602.01783",
}

@Article{florian,
  title =	"Reinforcement Learning Through Modulation of
		 Spike-Timing-Dependent Synaptic Plasticity",
  author =	"Razvan V. Florian",
  journal =	"Neural Computation",
  year = 	"2007",
  number =	"6",
  volume =	"19",
  bibdate =	"2018-11-14",
  bibsource =	"DBLP,
		 http://dblp.uni-trier.de/https://doi.org/10.1162/neco.2007.19.6.1468;
		 DBLP,
		 http://dblp.uni-trier.de/https://www.wikidata.org/entity/Q47841509;
		 DBLP,
		 http://dblp.uni-trier.de/db/journals/neco/neco19.html#Florian07",
  pages =	"1468--1502",
}

@InProceedings{mass,
  author =	"Wolfgang Maass",
  title =	"On the Computational Power of Noisy Spiking Neurons",
  pages =	"211--217",
  booktitle =	"Advances in Neural Information Processing Systems",
  editor =	"David S. Touretzky and Michael C. Mozer and Michael E.
		 Hasselmo",
  volume =	"8",
  year = 	"1996",
  publisher =	"The {MIT} Press",
}

@article{seung,
  author = {Seung, H. S.},
  description = {CCNLab BibTeX},
  journal = {Neuron},
  number = 6,
  pages = {1063-1073},
  title = {Learning in Spiking Neural Networks by Reinforcement of Stochastic
	Synaptic Transmission},
  volume = 40,
  year = 2003
}

@Misc{hedonistic,
  title =	"The Hedonistic Neuron",
  author =	"A. H. Klopf",
  year = 1982,

}
\bibliographystyle{apsr} 



\end{document}